\documentclass{article}
\usepackage{amsmath}
\usepackage[utf8]{inputenc}
\usepackage{hyperref}
\usepackage[capitalize,nameinlink]{cleveref}
\usepackage{graphicx}
\usepackage{xcolor}
\usepackage{url}
\usepackage{tablefootnote}
\usepackage{float}
\usepackage{subcaption}

\DeclareMathOperator{\Att}{Att}
\DeclareMathOperator{\RNN}{RNN}
\DeclareMathOperator{\softmax}{softmax}

\title{Learning semantic sentence representations from visually grounded language without lexical knowledge}
\author
       {D. Merkx and S. L. Frank\\
       Centre for Language Studies,\\
       Radboud University, Nijmegen, the Netherlands
}

\begin{document}
\label{firstpage}
\maketitle

\begin{abstract}
Current approaches to learning semantic representations of sentences often use prior word-level knowledge. The current study aims to leverage visual information in order to capture sentence level semantics without the need for word embeddings. We use a multimodal sentence encoder trained on a corpus of images with matching text captions to produce visually grounded sentence embeddings. Deep Neural Networks are trained to map the two modalities to a common embedding space such that for an image the corresponding caption can be retrieved and vice versa. We show that our model achieves results comparable to the current state-of-the-art on two popular image-caption retrieval benchmark data sets: MSCOCO and Flickr8k. We evaluate the semantic content of the resulting sentence embeddings using the data from the Semantic Textual Similarity benchmark task and show that the multimodal embeddings correlate well with human semantic similarity judgements. The system achieves state-of-the-art results on several of these benchmarks, which shows that a system trained solely on multimodal data, without assuming any word representations, is able to capture sentence level semantics. Importantly, this result shows that we do not need prior knowledge of lexical level semantics in order to model sentence level semantics. These findings demonstrate the importance of visual information in semantics.
\end{abstract}

\section{Introduction}

Distributional semantics, the idea that words that occur in similar contexts have similar meanings, has been around for quite a while (e.g., \cite{Rubenstein65, Deerwester1990}). \cite{Rubenstein65} already studied ``how the proportion of words common to contexts containing word A and to contexts containing word B was related to the degree to which A and B were similar in meaning" (p. 627). State-of-the-art word embedding methods such as Word2Vec \cite{Mikolov2013} and GloVe \cite{Pennington2014} have shown meaningful clusters, correlations with human similarity judgements \cite{DeDeyne2017}, and have become widely used features that boost performance in several natural language processing (NLP) tasks such as machine translation \cite{Qi2018}. With the success of word embeddings, researchers are looking for ways to capture the meaning of larger spans of text, such as sentences, paragraphs, and even entire documents. Much less is known about how to approach this problem and early solutions tried to adapt word embedding methods to larger spans of text, for example, Skip-Thought sentence embeddings  \cite{kiros2015}, FastSent \cite{Hill2016}, and Paragraph-Vector \cite{Le2014}, which are related to the Skip-Gram word model by \cite{Mikolov2013}. Recently, there have also been successful sentence encoder models which are trained on a supervised task and then transferred to other tasks (e.g. \cite{Conneau2017, Yang2018, Kiela2018}).

So far, existing sentence embedding methods often require (pretrained) word embeddings \cite{Conneau2017, Kiela2018}, large amounts of data \cite{Hill2016}, or both \cite{DeBoom2016, Yang2018}. While word embeddings are successful at enhancing sentence embeddings, they are not very plausible as a model of human language learning. Firstly, a model using word embeddings makes the assumption that the words in its lexicon are the linguistic units bearing meaning. Secondly, these models assume that the process of language acquisition begins with lexical level knowledge before learning how to process longer utterances. Both of these assumptions are questionable.

\cite{Tomasello}, a proponent of usage-based models of language, argues that children learn many relatively fixed expressions (e.g., `how-are-you-doing') as single linguistic units. Furthermore, he argues that the linguistic units that children operate on early in language acquisition are entire utterances, before their language use becomes more adult-like. Indeed, research shows that in young children, much of their language use is constrained to (parts of) utterances they have used before \cite{lieven2003} or comes from a small set of patterns like: `Where is X' and `Want more X' \cite{Braine}. Children's linguistic units become smaller and more adult-like as they learn to identify slots in the linguistic patterns and learn which constituents of their linguistic units they can `cut and paste' to create novel utterances \cite{Pine1993, Tomasello}. Models that assume lexical items are the basic meaning bearing units and that language learning starts from lexical items towards understanding full sentences are thus not very plausible as models of language learning. 

In the current study, we train a sentence encoder without prior knowledge of lexical semantics, that is, without using word embeddings. Instead of  word embeddings, we use character level input in conjunction with visual features. The use of multimodal data has proven successful on the level of word embeddings (see for instance \cite{Collell2017, Derby2013}). For sentence semantics, the multimodal task of image-caption retrieval, where given a caption the model must return the matching image and vice versa, has been proposed as a way of grounding sentence representations in vision \cite{Harwath2015, Leidal2018}. Recently \cite{Kiela2018} found that such models do indeed produce embeddings that are useful in tasks like natural language inference, sentiment analysis and subjectivity/objectivity classification. 

Our model does not know a priori which constituents of the input are important. It may learn to extract features from spans of text both larger and smaller than words. Furthermore, we leverage the potential semantic information that can be gained from the visual features to create visually grounded sentence embeddings without the use of prior lexical level knowledge. We also probe the semantic content of the grounded sentence embeddings more directly than has so far been done, by evaluating on Semantic Textual Similarity, a well known benchmark test set consisting of sentence pairs with human-annotated semantic similarity ratings.

Our aim is to create a language model that learns semantic representations of sentences in a more cognitively plausible way, that is, not purely text based and without prior lexical level knowledge. We evaluate our multi-modal sentence encoder on a large benchmark of human semantic similarity judgements in order to test if the similarity between the embeddings correlates with human judgements of semantic textual similarity. This is to the best of our knowledge the first evaluation of the sentence level semantics of a multimodal encoder that does not make use of lexical information in the form of word embeddings. We find that the model produces sentence embeddings that account for human similarity judgements, with performance similar to competing models. Importantly, our model does so using visual information rather than prior knowledge such as word embeddings. We release the code of our preprocessing pipeline, models and evaluation on github as open source: \url{https://github.com/DannyMerkx/caption2image}. 

\section{Sentence embeddings}

\subsection{Text-only methods}

Methods for creating sentence embeddings have thus far mostly been based solely on text data. Skip-Thought \cite{kiros2015}, inspired by the idea behind word embeddings, assumes that sentences which occur in similar context have similar meaning. Skip-Thought encodes a sentence and tries to reconstruct the previous sentence and the next sentence from the resulting embedding. In a similar approach, \cite{Yang2018} try to match Reddit posts with their responses based on the assumption that posts with similar meanings will elicit similar responses.

InferSent, a recent model by \cite{Conneau2017}, is one of the most successful models with regards to transfer learning and semantic content. \cite{Conneau2017} trained an RNN sentence encoder on the Stanford Natural Language Inference database \cite{snli:emnlp2015}, a database with paired sentences annotated for entailment, neutral, or contradiction relationships. \cite{Conneau2018} released SentEval, a transfer learning evaluation toolbox for sentence embeddings, which includes a large number of human semantic similarity judgements. InferSent embeddings show a high correlation to several sets of semantic textual similarity judgements and perform well on various transfer tasks like sentiment analysis and subjectivity/objectivity detection. 

\subsection{Multimodal methods}

Image-caption retrieval is a multimodal machine learning task involving challenges from both computer vision and language modelling. The task is to rank captions by relevance to a query image, or to rank images by relevance to a query caption, which is done by mapping the images and captions to a common embedding space and minimising the distance between the image and caption in this space.

\cite{Ma2015} used two Convolutional Neural Networks (CNN) to create image and sentence representations and another CNN followed by a Multilayer Perceptron (MLP) to derive a matching score between the images and captions. \cite{Klein2015} converted the captions to Fisher vectors \cite{Jaakkola1999} and used Canonical Correlations Analysis to map the caption and image representations to a common space. The model by \cite{Karpathy2017} works at a different granularity: They encoded image regions selected by an object detection CNN and encoded each word in the sentence separately, thus ending up with multiple embeddings per caption and image. They then calculated the distances between all the embedded words and image regions. 

Many image-caption retrieval models rely on pretrained neural networks and word embeddings. It is common practice to use a pretrained network such as VGG, Inception V2, or ResNet-152 to extract the visual features (e.g., \cite{Ma2015, Vendrov2015, Faghri2018, Wehrmann2018, Kiela2018}). Furthermore, with the exception of the character based model by \cite{Wehrmann2018}, recent results are achieved by using pretrained Word2Vec or GloVe word embeddings to initialise the sentence encoder. The current state-of-the-art results are by \cite{Faghri2018}, who fine-tuned a pretrained ResNet-152 and improved the sampling of mismatched image-caption pairs during training.  

The approach of mapping the image-caption pairs to a common semantic embedding space is interesting because the produced embeddings could also be useful in other tasks, similar to how word embeddings can be useful in machine translation \cite{Qi2018}. \cite{Kiela2018} used a model similar to \cite{Dong2018}, that is, a recurrent neural network caption encoder paired with a pretrained image recognition network which is trained to map the caption to the image features extracted by the image recognition network. Using SentEval, \cite{Kiela2018} showed that the resulting embeddings are useful in a wide variety of transfer tasks such as sentiment analysis in product and movie reviews, paraphrase detection and natural language inference. These results show that visually grounded sentence representations can be used for transfer learning, but do not directly probe the model's ability to learn sentence semantics. 

The current study differs from previous research in three respects. Firstly, we train our model using character level input rather than word embeddings. Secondly, our model uses only the sentence representations that can be learned from the multimodal training data. In contrast, \cite{Kiela2018} augmented their grounded representations by combining them with non-grounded (Skip-Thought) representations. Finally, we probe the semantic content of our sentence representations more directly by evaluating the caption encoder on the Semantic Textual Similarity benchmark. This benchmark is included in the SentEval toolbox but has to the best of our knowledge not been used to evaluate visually grounded sentence representations.

\section{Approach}

In this section, we first describe our encoder architectures, where we combine several best practices and state-of-the-art methods in the field of deep learning. Next, we describe the training data and finally the semantic similarity tasks. 

\subsection{Encoder Architectures}

\subsubsection{Image encoder}

\begin{figure}
    \centering
    \includegraphics[width =\textwidth]{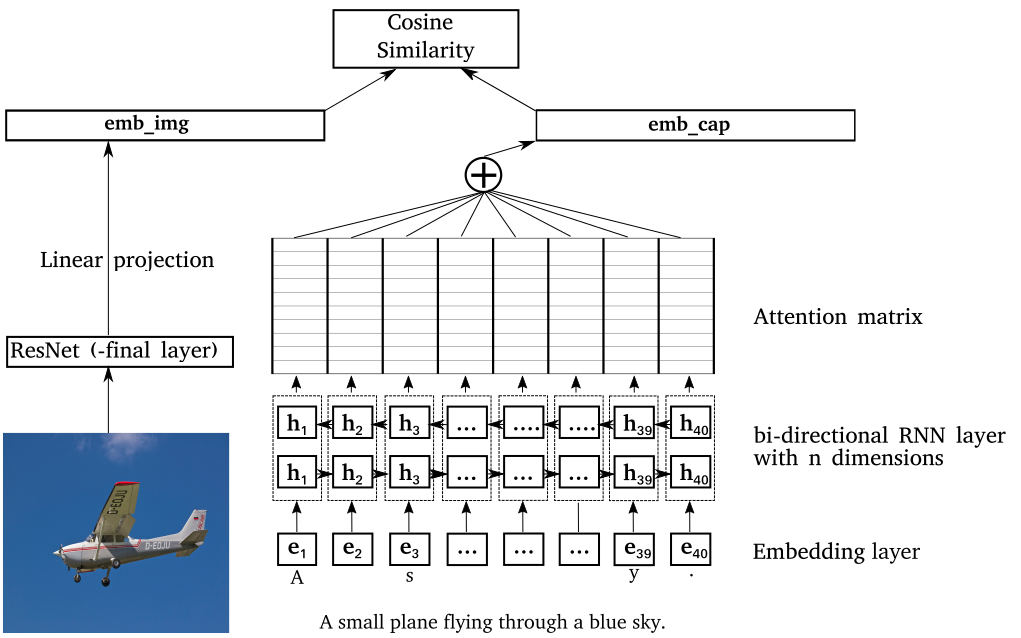}
    \caption{Model architecture: The model consists of two branches with the image encoder on the left and the caption encoder on the right. The character embeddings are denoted by $\mathbf{e}_t$ and the RNN hidden states by $\mathbf{h}_t$. Each hidden state has $n$ features which are concatenated for the forward and backward RNN into $2n$ dimensional hidden states. Then attention is applied which weighs the hidden states and then sums over the hidden states resulting in the caption embedding. At the top we calculate the cosine similarity between the image and caption embedding ($\textbf{emb\_img}$ and $\textbf{emb\_cap}$).}
    \label{network}
\end{figure}

Our model maps images and corresponding captions to a joint embedding space, that is, the encoders are trained to make the embeddings of an image-caption pair lie close to each other in the embedding space. As such the model requires both an image encoder and a sentence encoder as illustrated in Figure \ref{network}. 

The image features are extracted by a pretrained image recognition model trained on ImageNet \cite{imagenet_cvpr09}. For this we used ResNet-152 \cite{He2015}, a residualised network with 152 layers from which we take the activations of the penultimate fully connected layer\footnote{The final layer of a pretrained visual network is a task-specific object classification layer while the penultimate layer contains generally useful image features. This is backed up by research showing that the features of the penultimate layer yield better transfer learning results than the object classification layer \cite{Madhyastha2018}.}. ResNet-152 has lower error rates on the ImageNet task than other networks previously used in the image captioning task such as VGG16, VGG19 and Inception V2. 

For the image encoder we use a single layer linear projection on top of the pretrained image recognition model, and normalise the result to have unit L2 norm:
\[\mathbf{emb\_img} = \frac{\mathbf{img}A^T+\mathbf{b}}
{||\mathbf{img}A^T+\mathbf{b}||_2}\]
where $A$ and $\mathbf{b}$ are learned weights and bias terms, and $\mathbf{img}$ is the vector of ResNet image features. 

\subsubsection{Caption encoder}

We built a caption encoder that trains on raw text, that is, character-level input. The sentence encoder starts with an embedding layer with embeddings $(\mathbf{e}_1, ..., \mathbf{e}_t)$ for the $t$ characters in the input sentence. The embeddings are then fed into an RNN, followed by a self-attention layer and lastly normalised to have unit L2 norm:

\[\mathbf{emb\_cap} = \frac{\Att(\RNN(\mathbf{e}_1, ..., \mathbf{e}_t))}{||\Att(\RNN(\mathbf{e}_1, ..., \mathbf{e}_t))||_2}\] 
where $\mathbf{e}_1, ..., \mathbf{e}_t$ indicates the caption represented as character embeddings and $\Att$ is the attention layer. The character embedding features are learned along with the rest of the network.

The RNN layer allows the network to capture long-range dependencies in the captions. Furthermore, by making the layer bidirectional we let the network process the captions from left to right and vice versa, allowing the model to capture dependencies in both directions. We then concatenate the results to create a single embedding. We test two types of RNN: the Long Short Term Memory unit (LSTM; \cite{hochreiter1997long}) and the Gated Recurrent Unit (GRU; see \cite{Greff2017} and \cite{Chung2014} for detailed descriptions of these RNNs). The GRU is a recurrent layer that is widely used in sequence modelling (e.g., \cite{Zhu2015, Patel2016, Conneau2017}). The GRU requires fewer parameters than the LSTM while achieving comparable results or even outperforming LSTMs in many cases \cite{Chung2014}. On the other hand, \cite{Conneau2017} found that an LSTM not only performed better than a GRU on their training task, but also generalised better to other tasks including semantic similarity. We test both architectures as it is not clear which is better suited for the image-captioning task. 

The self-attention layer computes a weighted sum over all the hidden RNN states: 
\[\begin{aligned}&\mathbf{a}_t = \softmax(V\tanh(W\mathbf{h}_t +\mathbf{b}_w)+\mathbf{b}_v) \\
&\Att(\mathbf{h}_1, ..., \mathbf{h}_t) = \sum\limits_{t}\mathbf{a}_t\circ\mathbf{h}_t\end{aligned}\]
where $\mathbf{a}_t$ is the attention vector for hidden state $\mathbf{h}_t$ and $W$, $V$, $\mathbf{b}_w$, and $\mathbf{b}_v$ indicate the weights and biases. The applied attention is then the sum over the Hadamard product between all hidden states $(\mathbf{h}_1, ..., \mathbf{h}_t)$ and their attention vector. 

While attention is part of many state-of-the-art NLP systems, \cite{Conneau2017} found that attention caused their model to overfit on their training task, giving worse results on transfer tasks. As a simpler alternative to attention, we also test max pooling, where we take for each feature the maximum value over the hidden states. 

Both encoders are jointly trained to embed the images and captions such that the cosine similarity between image and caption pairs is larger (by a certain margin) than the similarity between mismatching pairs, minimising the so-called hinge loss. The network is trained on a minibatch $B$ of correct image-caption pairs $(cap,img)$ where all other image-caption pairs in the minibatch serve to create counterexamples $(cap,img')$ and $(cap',img)$. We calculate the cosine similarity $\cos(x,y)$ between each embedded image-caption pair and subtract the similarity of the mismatched pairs from the matching pairs such that the loss is only zero when the matching pair is more similar by a margin $\alpha$. The hinge loss $L$ as a function of the network parameters \(\theta\) is given by: 

\[\begin{aligned}L(\theta) = \sum\limits_{(cap,img),(cap',img')\in B} \biggl(\max(0, \cos(cap,img') - \cos(cap,img) + \alpha) +\\
\max(0, \cos(img,cap') - \cos(img,cap) + \alpha)\biggr) \end{aligned} \]
where $(cap,img)\neq(cap',img')$.

\subsection{Training Data}

The multimodal embedding approach requires paired captions and images for which we use two popular image-caption retrieval benchmark datasets: Flickr8k \cite{Hodosh2015} and MSCOCO \cite{Chen2015}.

\subsubsection{Flickr8k}

Flickr8k is a corpus of 8,000 images taken from the online photo sharing application Flickr.com. Each image has five captions created using Amazon Mechanical Turk (AMT) where workers were asked to ``write sentences that describe the depicted scenes, situations, events and entities (people, animals, other objects)'' \cite[p. 860]{Hodosh2015}. We used the data split provided by \cite{Karpathy2017}, with 6,000 images for training and a development and test set of 1,000 images each. 

To extract the image features, all images are resized such that the smallest side is 256 pixels while keeping the aspect ratio intact. We take ten $224\times 224$ crops of the image: one from each corner, one from the middle and the same five crops for the mirrored image. We use ResNet-152 pretrained on ImageNet to extract visual features from these ten crops and then average the features of the ten crops into a single vector with 2,048 features. The character input is provided to the networks as is, including all punctuation and capitals.

\subsubsection{MSCOCO}

Microsoft Common Objects in Context (MSCOCO) is a large dataset of 123,287 images with five captions per image. The captions were gathered using AMT, with workers being asked to describe the important parts of the scene. Like \cite{Vendrov2015}, we use 113,287 images for training and 5,000 for development and testing each. The image and text features are extracted from the data following the same procedure used for Flickr8k. The only difference is that the captions are provided in a tokenised format and we create the character level input by concatenating the tokens with single spaces and adding a full stop to the end of each caption.

\subsection{Training procedure} 

The image-caption retrieval performance on the development set is used to tune the hyperparameters for each network. We found a margin $\alpha = 0.2$ for the loss function to work best on both the GRUs and LSTMs. Although performance was relatively stable in the range $0.15\leq\alpha\leq 0.25$, it quickly degraded outside this range. The networks were trained with a single layer bidirectional RNN and we tested hidden layer sizes $n \in \{512, 1024, 2048\}$. The number of hidden units determines the embedding size, which is $2n$ (due to the  RNN being bidirectional). The attention layer has 128 hidden units. The image encoder has $2n$ dimensions to match the size of the sentence embeddings. We use 20-dimensional character embeddings and found that varying the size of these embeddings has very little effect on performance.

The networks are trained using Adam \cite{Kingma2015} with a cyclic learning rate schedule based on \cite{Smith2015}. The learning rate schedule varies the learning rate $lr$ smoothly between a minimum and maximum bound ($lr_{\text{min}}$ and $lr_{\text{max}}$) over the course of four epochs as given by:
\[
lr = 0.5(lr_{\text{max}}-lr_{\text{min}})(1+\cos(\pi(1 + 0.5 step\times{mb}))) + lr_{\text{min}}
\]
where $step$ indicates the step size, that is, the number of minibatches for a full cycle of the learning rate, and $mb$ is the number of minibatches processed so far. We set the step size such that the learning rate cycle is four epochs. The cyclic learning rate has two advantages. Firstly, fine-tuning the learning rate can be a very time consuming process. \cite{Smith2015} found that the cyclic learning rate works within reasonable upper and lower bounds which are easy to find: simply set the upper and lower bound by selecting the highest and lowest learning rates for which the loss value decreases. Secondly, the learning rate schedule causes the network to visit several local minima during training, allowing us to use snapshot ensembling \cite{Huang2017}. By saving the network parameters at each local minimum, we can ensemble the caption embeddings of multiple networks at no extra cost.

We train the networks for 32 epochs and take a snapshot for ensembling at every fourth epoch. For ensembling we use the two snapshots with the highest performance on the development data. We found that for Flickr8k an upper bound on the learning rate of $10^{-3}$ and a lower bound of $10^{-6}$ worked well and for MSCOCO we had to adjust the upper bound to $10^{-4}$. 

\subsection{Semantic Evaluation}

For the semantic evaluation we use the SentEval toolbox introduced by \cite{Conneau2018}. This toolbox is meant to test sentence embeddings on a diverse set of transfer tasks, from sentiment analysis and paraphrase detection to entailment prediction. For semantic textual similarity analysis, SentEval includes the Semantic Textual Similarity and Sentences Involving Compositional Knowledge datasets which we briefly review here. After training our multimodal encoder network, we simply discard the image encoder, and the caption encoder is used to encode the test sentences in SentEval.  

\subsubsection{Semantic Textual Similarity}

Semantic Textual Similarity (STS) is a shared task hosted at the SemEval workshop. SentEval covers the STS datasets from 2012 to 2016. The datasets consist of paired sentences from various sources labelled by humans with a similarity score between zero (`the two sentences are completely dissimilar') and five (`the two sentences are completely equivalent, as they mean the same thing') for a total of five annotations per sentence pair (\cite[p. 254]{Agirre2015}, see also for a full description of the annotator instructions). The evaluation performed on the STS 2012 to 2016 tasks measures the correlation between the cosine similarity of the sentence embeddings and the human similarity judgements. 

\begin{table}
    \caption{Description of the various STS tasks and their subtasks. Some subtasks appear in multiple STS tasks, but consist of different sentence pairs drawn from the same source. The image description datasets are drawn from the PASCAL VOC-2008 dataset \cite{pascal-voc-2008} and so do not overlap with Flickr8k or MSCOCO.}
    \begin{minipage}{1.18\textwidth}
        \resizebox{.85\linewidth}{!}{
        \begin{tabular}{l l r l } 
        \hline\hline
         Task & Subtask & \#Pairs & Source \\ \hline
          & MSRpar & 750 & newswire\\
         & MSRvid & 750 & videos\\
         STS 2012 & SMTeuroparl & 459 & glosses\\
         & OnWN & 750 & WMT eval.\\
         & SMTnews & 399 & WMT eval.\\ \hline
         & FNWN & 189 & newswire\\
         STS 2013 & HDL & 750 & glosses\\
         & OnWN & 561 & glosses\\ \hline
         & Deft-forum & 450 & forum posts\\
         & Deft-news & 300 & news summary\\
         STS 2014 & HDL & 750 & newswire headlines\\
         & Images & 750 & image descriptions\\
         & OnWN & 750 & glosses\\
         & Tweet-news & 750 & tweet-news pairs\\ \hline
         & Answers forum & 375 & Q\&A forum answers\\
         & Answers students & 750 & student answers\\
         STS 2015 & Belief & 375 & committed belief\\
         & HDL & 750 & newswire headlines\\
         & Images & 750 & image descriptions\\ \hline
         & Answer-Answer & 254 & Q\&A forum answers\\
         & HDL & 249 & newswire headlines\\
         STS 2016 & Plagiarism & 230 & short-answer plagiarism\\
         & Postediting & 244 & MT postedits\\
         & Question-Question & 209 &  Q\&A forum questions\\ \hline
         Total & & 12,544 & \\\hline \hline
        \end{tabular}}
    \end{minipage}
    \label{STS}
\end{table}
The STS Benchmark set (STS-B) consists of 8,628 sentence pairs selected from all STS tasks \cite{Cer2017}. STS-B consists of a training, development and test set (5,749, 1,500 and 1,379 sentence pairs respectively). For the STS-B task, the SentEval toolbox trains a classifier which tries to predict the similarity scores using the sentence embeddings resulting from our model. Table \ref{STS} gives an overview of the datasets. For full descriptions of each dataset see \cite{Agirre2012, Agirre2013, Agirre2014, Agirre2015, Agirre2016}.

\subsubsection{Sentences Involving Compositional Knowledge}

Sentences Involving Compositional Knowledge (SICK) is a database created for a shared task at SemEval-2014 with the purpose of testing compositional distributional semantics models \cite{Bentivogli2016}. The dataset consists of 10,000 sentence pairs which were generated using sentences taken from Flickr8k and the STS 2012 MSRvid data set. The sentences were altered to display linguistic phenomena that the shared task was meant to evaluate, such as negation. This resulted in sentences like `there is no biker jumping in the air' and `two angels are making snow on the lying children' (altered from `two children are lying in the snow and are making snow angels', \cite[p. 6]{Bentivogli2016}) which do not occur in the Flickr8k training data. 

For the semantic evaluation of our sentence embeddings we used the SICK Relatedness (SICK-R) annotations. For the SICK-R task, annotators were asked to rate the relatedness of sentence pairs on a 5-point scale for a total of ten annotations per sentence pair. Unlike for STS, there were no specific descriptions attached to the scale; participants were only instructed using examples of related and unrelated sentence pairs. Similar to STS-B, a classifier is trained on top of the embeddings, using 45 percent of the data as training set, 5 percent as development set and 50 percent as test set.

\section{Results and Discussion}

\subsection{Model Selection}

We perform model selection after training on only the Flickr8k database. Due to the considerably larger size of MSCOCO it is more efficient to train and test our models on Flickr8k, and train on MSCOCO using only the best setup found on Flick8k.
\begin{figure}
    \begin{minipage}{1\textwidth}
        \centering
        \includegraphics[width =\textwidth]{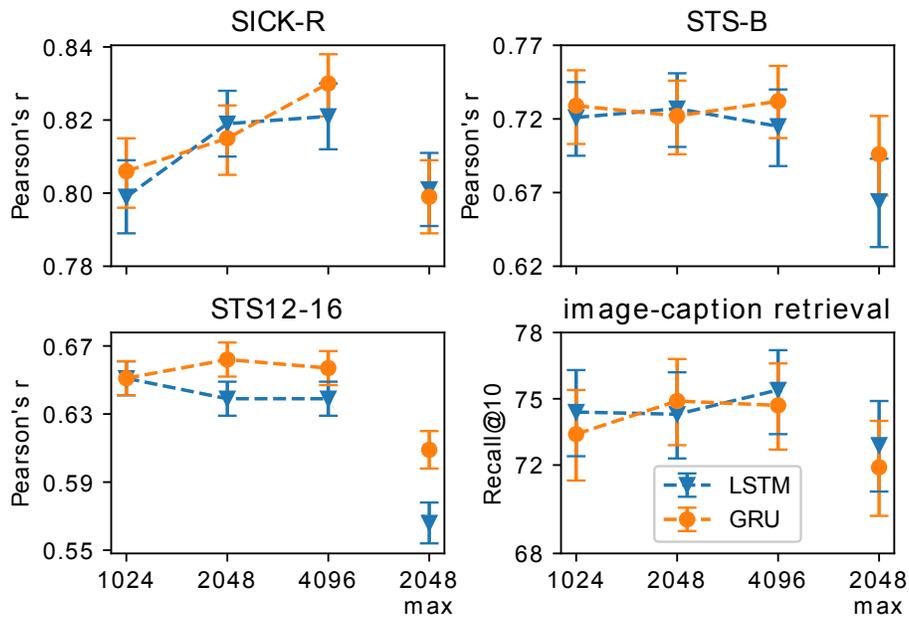}
        \caption{Model performance on the semantic (SICK-R, STS-B, and STS12-16) and training task (image-caption retrieval) measures including the 95 percent confidence interval. Training task performance is measured in recall@10. The semantic performance measure is Pearson's $r$. The horizontal axis shows the embedding size with max indicating the max pooling model.}
        \label{barplots}
    \end{minipage}
\end{figure}

To select the DNN architecture with the best performance we compare our architectures on image-caption retrieval performance and on their ability to capture semantic content. The image-caption retrieval performance is measured by Recall@10: the percentage of images (or captions) for which the correct caption (or image) was in the top ten retrieved items. For the purpose of model selection we use the average of the bidirectional (caption to image and image to caption) retrieval results on the development set. For the semantic evaluation we use correlation coefficients (Pearson's $r$) between embedding distances and human similarity judgements from STS-B and SICK-R. We also aggregate the Pearson's $r$ scores for the STS 2012 through 2016 tasks. 

Figure \ref{barplots} shows the results for our models trained on Flickr8k.
There is no clear winner in terms of performance: The GRU 2048 (referring to the embedding size) performs best on STS, GRU 4096 on SICK-R and STS-B, and LSTM 4096 on the training task. Although there are differences between the GRU and the LSTM, they are only statistically significant for STS12-16. Furthermore, the max pooling models are outperformed by their attention based counterparts. We only tested the max pooling with an embedding size of 2048. Due to the clear drop in both training and semantic task performance we did not run any further experiments.

As our main goal is the evaluation of semantic content, we continue with the GRUs as they perform significantly better on STS12-16. There is no clear winner between the GRU 2048 and GRU 4096 as the performance differences on all measures are relatively small. The 4096 model performs significantly better on SICK-R but the 2048 model performs slightly better on STS12-16. As STS12-16 is the main interest in our evaluation we pick the GRU 2048 as our best performing Flickr8k model and train a GRU 2048 model on MSCOCO. We will from now on refer to this model as char-GRU, shorthand for character based GRU.

\subsection{Image-Caption retrieval}

\begin{table}
    \caption{Image-Caption retrieval results on the Flickr8k test set. R@N is the percentage of items for which the correct image or caption was retrieved in the top N (higher is better). Med r is the median rank of the correct image or caption (lower is better). We also report the 95 percent confidence interval for the R@N scores.}
    \begin{minipage}{1.45\textwidth}
        \resizebox{.7\linewidth}{!}{
        \begin{tabular}{l | r r r r | r r r r}
        \hline\hline
            \multicolumn{1}{l}{Model} & \multicolumn{4}{c}{Caption to Image} & \multicolumn{4}{c}{Image to Caption}  \\ \hline
             & R@1 & R@5 & R@10 & med r & R@1 & R@5 & R@10 & med r \\
            \cite{Karpathy2017} & 11.8$\pm$2.0 & 32.1$\pm$2.9 & 44.7$\pm$3.1 & 12.4 & 16.5$\pm$2.3 & 40.6$\pm$3.0 & 54.2$\pm$3.1 & 7.6 \\
            \cite{Ma2015} & 20.3$\pm$2.5 & 47.6$\pm$3.1 & 61.7$\pm$3.0 & 5.0 & 24.8$\pm$2.7 & 53.7$\pm$3.1 & 67.1$\pm$2.9 & 5.0 \\
            \cite{Klein2015} & 21.2$\pm$2.5 & 50.0$\pm$3.1 & 64.8$\pm$3.0 & 5.0 & 31.0$\pm$2.9 & 59.3$\pm$3.0 & 73.7$\pm$2.7 & 4.0 \\
            \cite{Wehrmann2018} & 26.9$\pm$2.7 & - & 69.6$\pm$2.9 & 4.0 & 32.4$\pm$2.9 & - & 73.6$\pm$2.7 &  3.0  \\
            \cite{Dong2018} & - & - & - & - & 36.3$\pm$3.0 & 66.4$\pm2.9$ & 78.2$\pm2.6$ & -\\ 
            char-GRU & 27.5$\pm$2.8 & 58.2$\pm$3.1 & 70.5$\pm$2.8 & 4.0 & 38.5$\pm$3.0 & 68.9$\pm$2.9 & 79.3$\pm$2.5 & 2.0\\ \hline\hline
        \end{tabular}}
    \end{minipage}
    \label{flickr_c2i_results}
\end{table}

\begin{table}
    \caption{Image-Caption retrieval results on the MSCOCO test set. We report the results on the full test set (5,000 items) and the average results on five folds of 1,000 image-caption pairs. R@N is the percentage of items for which the correct image or caption was retrieved in the top N (higher is better). Med r is the median rank of the correct image or caption (lower is better). We also report the 95 percent confidence interval for the R@N scores.}
    \begin{minipage}{1.45\textwidth}
        \resizebox{.7\linewidth}{!}{
        \begin{tabular}{l | r r r r | r r r r}
        \hline\hline
             \multicolumn{1}{l}{Model} & \multicolumn{4}{c}{Caption to Image} & \multicolumn{4}{c}{Image to Caption}  \\ \hline
             \multicolumn{1}{}{}& R@1 & R@5 & R@10 & med r & R@1 & R@5 & R@10 & med r \\ \hline
             \multicolumn{1}{}{}&\multicolumn{8}{c}{1k results}\\\hline
             \cite{Klein2015} & 25.1$\pm$1.2 & 59.8$\pm$1.4 & 76.6$\pm$1.2 & 4.0 & 39.4$\pm$1.4 & 67.9$\pm$1.3 & 80.9$\pm$1.1 & 2.0 \\
             \cite{Ma2015} & 32.6$\pm$1.3 & 68.6$\pm$1.3 & 82.8$\pm$1.0 & 3.0 & 42.8$\pm$1.4 & 73.1$\pm$1.2 & 84.1$\pm$1.0 & 2.0 \\
             \cite{Vendrov2015} & 37.9$\pm$1.3 & - & 85.9$\pm$1.0 & 2.0 & 46.7$\pm$1.4 & - & 88.9$\pm$0.9 & 2.0 \\
             \cite{Faghri2018} & 52.0$\pm$1.4 & 84.3$\pm$1.0 & 92.0$\pm$0.8 & 1.0 & 64.6$\pm$1.3 & 90.0$\pm$0.8 & 95.7$\pm$0.6 & 1.0 \\
             \cite{Wehrmann2018} & 40.4$\pm$1.4 & - & 88.6$\pm$0.9 & 2.0 & 49.5$\pm$1.4 & - & 91.3$\pm$0.8 & 1.6 \\
             char-GRU & 41.4$\pm$1.4& 76.8$\pm$1.2& 88.0$\pm$0.9& 2.0 & 51.2$\pm$1.4& 83.5$\pm$1.0& 92.1$\pm$0.7& 1.2 \\ \hline
             \multicolumn{1}{}{}& \multicolumn{8}{c}{5k results}\\ \hline
             \cite{Klein2015} & 10.8$\pm$0.9 & 28.3$\pm$1.2 & 40.1$\pm$1.4 & 17.0 & 17.3$\pm$1.0 & 39.0$\pm$1.4 & 50.2$\pm$1.4 & 10.0 \\
             \cite{Vendrov2015} & 18.0$\pm$1.1 & - & 57.6$\pm$1.4 & 7.0 & 23.3$\pm$1.2 & - & 65.0$\pm$1.3 & 5.0 \\ 
             \cite{Faghri2018} & 30.3$\pm$1.3 & 59.4$\pm$1.4 & 72.4$\pm$1.2 & 4.0 & 41.3$\pm$1.4 & 71.1$\pm$1.3 & 81.2$\pm$1.1 & 2.0 \\
             \cite{Kiela2018} & 17.1$\pm$1.0 & 43.0$\pm$1.4 & 57.3$\pm$1.4 & 8.0 & 27.1$\pm$1.2 & 55.6$\pm$1.4 & 70.0$\pm$1.3 & 4.0 \\
             char-GRU & 20.2$\pm1.1$& 46.9$\pm1.4$& 60.9$\pm1.4$& 6.0& 25.7$\pm1.2$& 54.3$\pm1.4$& 68.8$\pm1.3$& 4.0 \\ \hline\hline
        \end{tabular}}
    \end{minipage}
    \label{coco_c2i_results}
\end{table}

We compare our char-GRU model with the current state-of-the-art in image-caption retrieval on both Flickr8k and MSCOCO. Tables \ref{flickr_c2i_results} and \ref{coco_c2i_results} show the bidirectional retrieval results on Flickr8k and MSCOCO, respectively. For MSCOCO we report both the results on the full test set (5000 items) and average results on a five-fold test set of 1000 items to be able to compare our results to previous work. Our models perform comparable to the state-of-the-art on both image to caption and caption to image retrieval on all metrics for Flick8k. The MSCOCO model by \cite{Faghri2018}, which fine-tuned the ResNet-152 network during training, is the only model that significantly outperforms our own across the board. 

All systems except the one by \cite{Wehrmann2018} and our own made use of word embeddings. \cite{Wehrmann2018} report that their CNN model trained on Flickr8k could only achieve such high recall scores when fine-tuning a model that was pretrained on MSCOCO, which they hypothesised is due to the small number of training examples in Flickr8k. Using our char-GRU model we outperform their convolutional approach without any pretraining on MSCOCO, indicating that Flickr8k has enough training examples for a recurrent architecture to take advantage of.

\subsection{Semantic Evaluation}

\begin{figure}
    \begin{minipage}{1\textwidth}
        \centering
        \includegraphics[width =\textwidth]{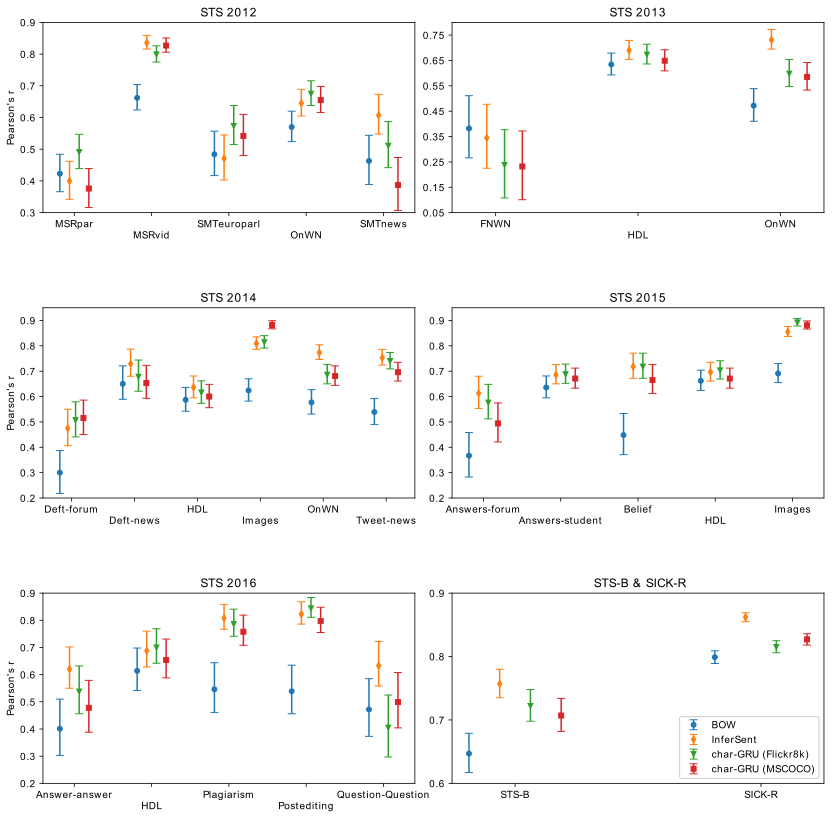}
        \caption{Semantic evaluation task results: Pearson correlation coefficients with their 95 percent confidence interval for the various subtasks (see Table~\ref{STS}). BOW is a bag of words approach using GloVe embeddings and InferSent is the model reported by \cite{Conneau2017}. Appendix A contains a table of the results shown here.}
        \label{errorplots}
    \end{minipage}
\end{figure}

We now look at the semantic properties of the sentence embeddings in more detail and compare our models with previous work. Figure \ref{errorplots} displays Pearson's $r$ scores on all the subtasks of the STS tasks for our char-GRU model, InferSent \cite{Conneau2017}, and a Bag Of Words (BOW) baseline using the average over a sentence's GloVe vectors. 

\subsubsection{Comparing Flickr8k with MSCOCO}

First of all, our Flickr8k model significantly outperforms the MSCOCO model on 6 out of 26 tasks, while the MSCOCO model only outperforms the Flickr8k model on MSRvid, Images (STS 2014) and SICK-R. It seems that the larger amount of image-caption data in MSCOCO allows the model to become better at what it was already good at, that is, video and image descriptions. On the other hand, specialising in image and video descriptions seems to decrease the models' generalisation to other tasks indicating that it is overfitting. That being said, the Flickr8k model performs quite well, beating the InferSent and BOW models on some tasks and performing comparably on most of the other tasks even though the Flickr8k database is only about five percent of the size of MSCOCO and about one percent of what InferSent is trained on. 

\subsubsection{Comparing with BOW baseline}

It is important to note that models using GloVe vectors receive a considerable amount of prior lexical semantic knowledge. GloVe vectors are trained on an 840-billion-word corpus with a vocabulary of over 2.2 million words and InferSent gets all of this extracted semantic knowledge for free. If the model encounters a word in the transfer tasks that it has never seen during training, it still has knowledge of the word's semantic relatedness to other words through that word's GloVe vector. 

This makes the BOW model a useful baseline model. It uses the prior word knowledge that InferSent uses (GloVe vectors) but it is not trained to create sentence embeddings. While InferSent is a significant improvement over the BOW model on most tasks (22 out of 26), it does not improve on the BOW model on 4 out of 26 tasks. Figure \ref{errorplots} shows that the BOW model performs close to the three trained models on many tasks. InferSent and the BOW model have the same input, but InferSent is trained on large amounts of data in order to extract information from this input. This then makes it reasonable to assume that a large part of InferSent's performance is due to the word level semantic information available in the GloVe vectors. 

Our char-GRU model does not have such information available but instead benefits from being grounded in vision. By learning language from the ground up from multimodal data, our model learns to capture sentence semantics with a performance comparable to models which receive prior knowledge of lexical semantics. Even though the system's only language input consists of image captions, Figure \ref{errorplots} shows that our model generalises well to a wide variety of domains. The Flickr8k model significantly outperforms the BOW baseline on 20 out of 26 tasks. 

\subsubsection{Comparing with InferSent}

Next, we compare InferSent with our Flickr8k char-GRU in more detail. Our model performs on par with InferSent on 16 out of 26 tasks. It is not surprising that our char-GRU model performs well on the Images sets, with a significant improvement over InferSent on Images (STS 2015). Our char-GRU also outperforms InferSent significantly by quite a margin on SMTeuroparl (transcriptions from European Parliament sessions) and MSRpar (a news set scraped from the internet), both very different from each other and different from image captions. Table \ref{examples} contains examples of these datasets to highlight what we will discuss next.

On closer inspection, SMTeuroparl contains sentence pairs with high word overlap and relatively high similarity scores given by the human annotators. Even though word embedding based models should be just as capable of exploiting high word overlap as our char-GRU model, perhaps they are more prone to make mistakes if the two sentences differ by a very rare word such as `pontificate' in the example. The embedding for such a rare word could be very skewed towards an unrepresentative context when learning the embeddings. The MSRpar dataset contains many proper nouns for which no embedding might exist and it is common practice to then remove the word from the input. In contrast, our character based method does not remove such proper nouns and thereby benefits from morphological similarity between the two sentences, even though the proper noun has never been seen before. Indeed, our model seems to work reasonably well on the other news databases as well, achieving state-of-the-art performance equal to InferSent on all HDL (news headlines) sets. 

InferSent significantly outperforms our Flickr8k trained char-GRU model on 7 out of 26 tasks. Especially noticeable is our model's performance on the Question-Question (forum question) dataset and on FNWN (WordNet definitions), the only task where our model is outperformed significantly by the BOW model. FNWN contains definition-like sentences, often with structures that one does not find in an image description. In the example in Table \ref{examples}, for instance, the first sentence of the pair is very lengthy and contains parentheses and abbreviations, while the second sentence is very short and lacks a subject. Concerning the question database, our model has never seen a question during training. Questions have a different syntactic structure than what our model has seen during training. Furthermore, most image descriptions tend to start with the word `A' (e.g., `A man scales a rock in the forest.'), whereas questions tend to start with `What', `Should' and `How', for example. 

\begin{table}
    \caption{Example sentence pairs with their human-annotated similarity score taken from STS tasks. }
    \begin{minipage}{1.45\textwidth}
        \resizebox{.7\linewidth}{!}{
        \begin{tabular}{l c p{0.7\textwidth}} 
        \hline\hline
        Dataset & Similarity & Example pair \\ \hline
        SMTeuroparl & 3.5 & We often pontificate here about being the representatives of the citizens of Europe.\\ 
        &  & We are proud often here to represent the citizens of Europe.\\ \hline
        MSRpar & 4.6 & Myanmar's pro-democracy leader Aung San Suu Kyi will return home late Friday but will remain in detention after recovering from surgery at a Yangon hospital, her personal physician said. \\
        &  & Myanmar's pro-democracy leader Aung San Suu Kyi will be kept under house arrest following her release from a hospital where she underwent surgery, her personal physician said Friday. \\ \hline
        FNWN & 2.0 & An agent has attempted to achieve a goal, and the actual outcome of the agent's action has been resolved, so that it either specifically matches the agent's intent (e.g. success) or does not match it (e.g. failure). \\
        & & Having succeeded or being marked by a favorable outcome. \\ \hline
        Question-Question& 4.0 &  Should I drink water during my workout? \\
        & & How can I get my toddler to drink more water? \\ \hline \hline
        \end{tabular}}
    \end{minipage}
    \label{examples}
\end{table}

\subsubsection{Trade-off between training task and transfer task performance}

\begin{figure}
    \begin{minipage}{1\textwidth}
    
        \begin{subfigure}[b]{0.5\textwidth}
            \includegraphics[width=\textwidth]{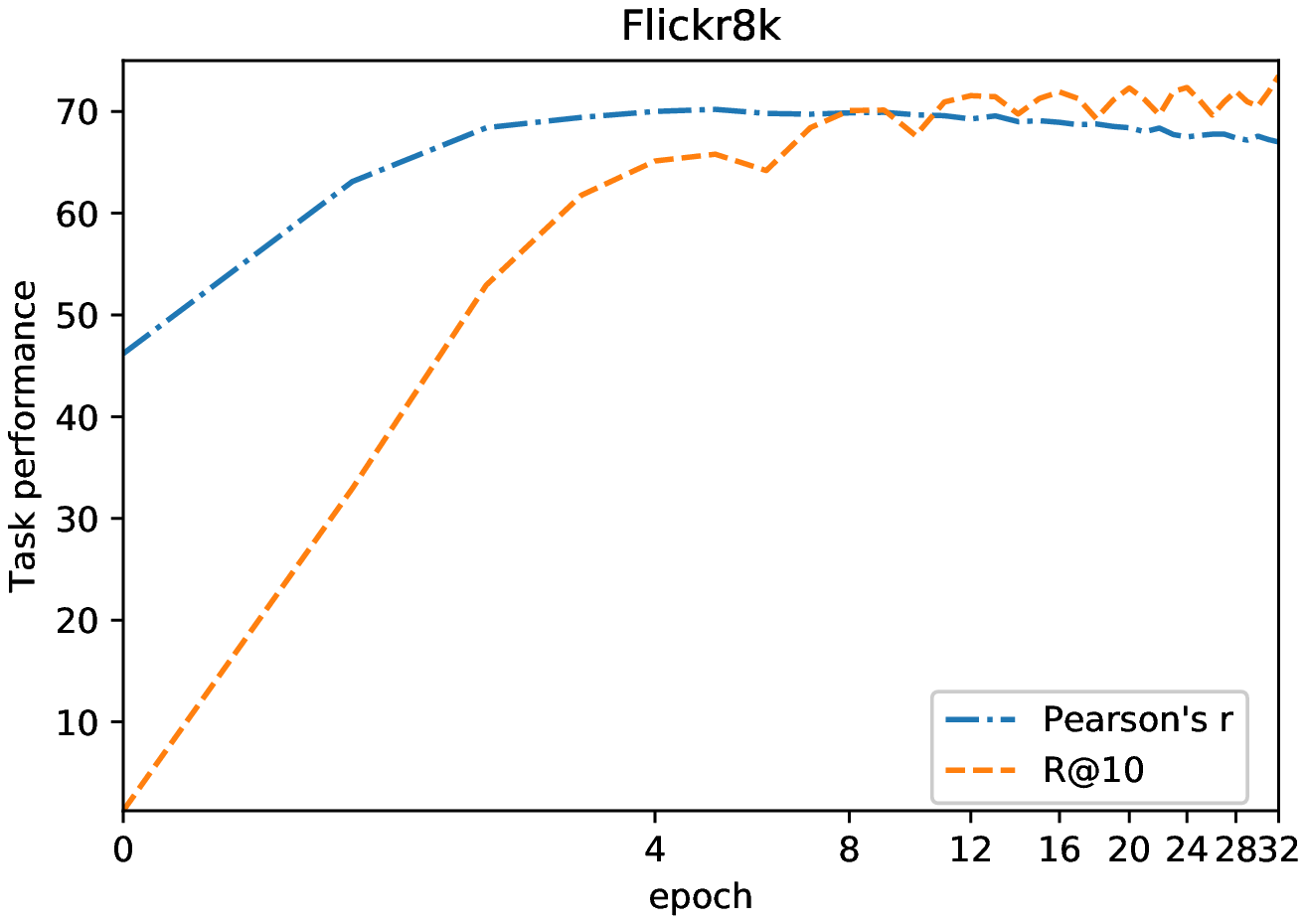}
        \end{subfigure}
        \begin{subfigure}[b]{0.5\textwidth}
            \includegraphics[width=\textwidth]{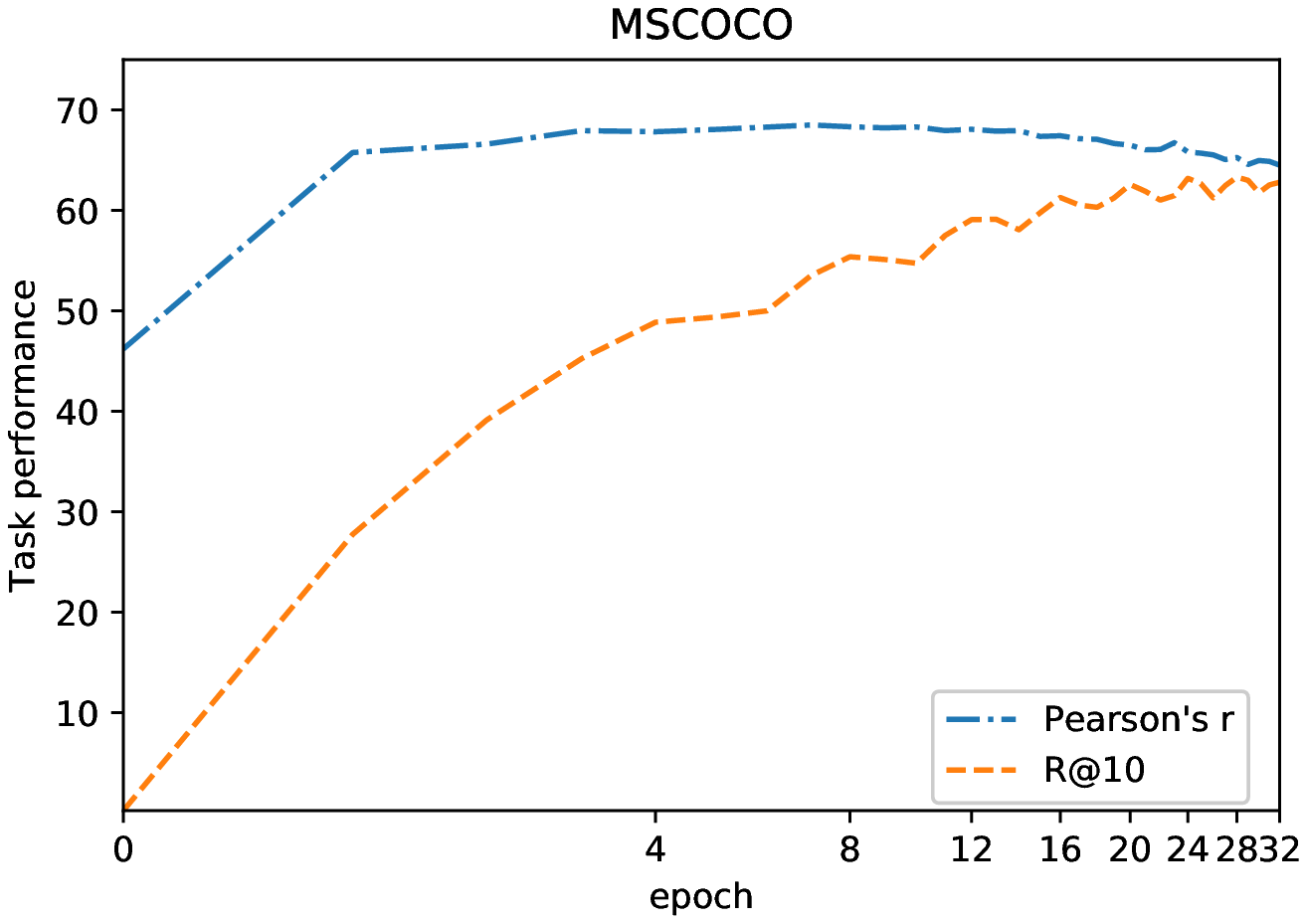}
        \end{subfigure}
        \caption{The training task performance (R@10) and the semantic task performance (Pearson's $r \times 100$) as they develop over training, with the number of epochs on a logarithmic scale. For MSCOCO (right) we show the training task performance on the 5,000 item test set.}
        \label{pearsonplots}
    \end{minipage}
\end{figure}

We further investigate how prone our model is to overspecialising on image descriptions. Figure \ref{pearsonplots} shows how the bidirectional image-caption retrieval performance and the semantic task performance (SICK-R and STS12-16 combined) develop during training. 

Epoch zero is the performance of an untrained model, and it is clear that both measures increase substantially during the first few epochs. Most improvement in both training task and semantic task performance happens in the first four epochs. After that the training task performance still increases by 12.8 and 28.5 percent for Flickr8k and MSCOCO, respectively. On the other hand, semantic task performance peaks around epoch four and then slowly decreases by 4.6 and 5.8 percent towards the last epoch for Flickr8k and MSCOCO, respectively. So even though our model is capable of learning how to extract semantic information from image-caption pairs, it is prone to overspecialising on the training task. The performance drop on the semantic task is only small, but trade-offs between the performance on different tasks poses a challenge to the search for universal sentence embeddings.

\section{Conclusion}

We investigated whether sentence semantics can be captured in sentence embeddings without using (prior) lexical knowledge. We did this using a multimodal encoder which grounds language in vision using image-caption pairs. \cite{Harwath2015} have claimed that this method produces a multimodal semantic embedding space and, indeed, we found that the distances between resulting sentence embeddings correlate well with human semantic similarity judgements, in some cases more so than models based on word embeddings. Importantly, this shows that we do not need to use word embeddings, which has hitherto been the standard in sentence embedding methods. The addition of visual information during training allows our model to capture semantic information from character-level language input. 

The model generalises well to linguistic domains such as European Parliament transcriptions, which are very different from the image descriptions it was trained on, but our model also has difficulty with some of the subtasks. For instance, our model scored significantly lower than InferSent on the SICK and forum question databases suggesting that our grounding approach alone is not enough to learn semantics for all linguistic domains. This could be because some visual information is hardly ever explicitly written down (few people will write down obvious facts like `bananas are yellow'), while more abstract concepts will not appear in images or their descriptions (e.g., the words `intent' and `attempted' from our test sentences in Table \ref{examples} are hard to capture in image). Future work could combine the visual grounding approach with text-only methods in order to learn from more diverse data. In such a multitask learning setting, our grounded sentence encoder could be fine-tuned on for instance natural language inference data, combining our approach with that of InferSent \cite{Conneau2017}.

In future work, we plan to work on spoken utterances. Unlike text, speech is not neatly segmented into lexical units, posing a challenge to conventional word embedding methods. However, the results presented here show that it is possible to learn sentence semantics without such prior lexical semantic knowledge and segmentation into lexical units. So far, studies of sentence meaning have mostly focused on written language, even though we learn to listen and speak long before we learn how to read and write. Learning representations of sentence meaning directly from speech therefore seems more intuitive than separately learning word and sentence representations from written sources. Furthermore, most languages have no orthography and only exist in spoken form. Capturing semantics directly from the speech signal provides a way to model sentence semantics for these languages. While there is previous work on spoken caption-image retrieval (e.g., \cite{Harwath2016, Chrupala2017}) we have barely scratched the surface of transfer learning using spoken input.

\section*{Acknowledgements}

The research presented here was funded by the Netherlands Organisation for Scientific Research (NWO) Gravitation Grant 024.001.006 to the Language in Interaction Consortium. This work was carried out on the Dutch national e-infrastructure with the support of SURF Cooperative. We would like to thank Mirjam Ernestus for commenting on an earlier version of this paper. 

\bibliographystyle{IEEEtran}

\newpage
\label{lastpage}
\pagebreak
\appendix

\section{Semantic evaluation results table}
Table~\ref{Semantic_results} shows the semantic evaluation results which were used to create Figure \ref{errorplots}.
\begin{table}[H]
    \caption{Semantic textual similarity results (Pearson's $r$ $\times100$ with 95 percent confidence interval). BOW is a bag of words approach using GloVe embeddings and InferSent is the model reported by \cite{Conneau2017}.}
    \begin{minipage}{1.2\textwidth}
        \resizebox{.85\linewidth}{!}{
        \renewcommand{\arraystretch}{1.1}
        \begin{tabular}{l l r r r r} 
        \hline\hline
              &         & \multicolumn{4}{c}{Model} \\
         Task & Dataset & BOW & InferSent & char-GRU & char-GRU \\
         & & & & (Flickr8k) & (MSCOCO)\\\hline
          & MSRpar & 42.3$\genfrac{}{}{0pt}{}{+5.7}{-6.1}$ & 40.0 $\genfrac{}{}{0pt}{}{+5.8}{-6.2}$ & 49.1$\genfrac{}{}{0pt}{}{+5.2}{-5.6}$ & 37.6$\genfrac{}{}{0pt}{}{+6.0}{-6.3}$\\
         & MSRvid & 66.2$\genfrac{}{}{0pt}{}{+3.8}{-4.2}$ & 83.6$\genfrac{}{}{0pt}{}{+2.0}{-2.3}$ & 79.9$\genfrac{}{}{0pt}{}{+2.4}{-2.7}$ & 82.7$\genfrac{}{}{0pt}{}{+2.1}{-2.4}$\\
         STS 2012 & SMTeuroparl & 48.4$\genfrac{}{}{0pt}{}{+6.7}{-7.3}$ & 47.1$\genfrac{}{}{0pt}{}{+6.8}{-7.4}$ & 57.3$\genfrac{}{}{0pt}{}{+5.8}{-6.5}$ & 54.2$\genfrac{}{}{0pt}{}{+6.2}{-6.8}$ \\
         & OnWN & 57.0$\genfrac{}{}{0pt}{}{+4.6}{-5.0}$ & 64.5$\genfrac{}{}{0pt}{}{+4.0}{-4.4}$ & 67.5$\genfrac{}{}{0pt}{}{+3.7}{-4.1}$ & 65.5$\genfrac{}{}{0pt}{}{+3.9}{-4.3}$\\
         & SMTnews & 46.3$\genfrac{}{}{0pt}{}{+7.4}{-8.1}$ & 60.7$\genfrac{}{}{0pt}{}{+5.9}{-6.6}$ & 51.1$\genfrac{}{}{0pt}{}{+6.9}{-7.6}$ & 38.7$\genfrac{}{}{0pt}{}{+8.0}{-8.7}$\\ \hline
         & FNWN & 38.2$\genfrac{}{}{0pt}{}{+11.6}{-12.9}$ & 34.5$\genfrac{}{}{0pt}{}{+12.0}{-13.2}$ & 23.8$\genfrac{}{}{0pt}{}{+13.0}{-13.9}$ & 23.2$\genfrac{}{}{0pt}{}{+13.1}{-14.0}$\\
         STS 2013 & HDL & 63.4$\genfrac{}{}{0pt}{}{+4.1}{-4.5}$ & 69.0$\genfrac{}{}{0pt}{}{+3.6}{-3.9}$ & 67.3$\genfrac{}{}{0pt}{}{+3.7}{-4.1}$ & 64.9$\genfrac{}{}{0pt}{}{+4.0}{-4.3}$\\
         & OnWN & 47.2$\genfrac{}{}{0pt}{}{+6.2}{-6.7}$ & 73.1$\genfrac{}{}{0pt}{}{+3.6}{-4.1}$ & 59.8$\genfrac{}{}{0pt}{}{+5.1}{-5.6}$ & 58.5$\genfrac{}{}{0pt}{}{+5.2}{-5.7}$\\\hline
         & Deft-forum & 30.0$\genfrac{}{}{0pt}{}{+8.2}{-8.7}$ & 47.5$\genfrac{}{}{0pt}{}{+6.9}{-7.5}$ & 50.7$\genfrac{}{}{0pt}{}{+6.6}{-7.2}$ & 51.5$\genfrac{}{}{0pt}{}{+6.5}{-7.1}$\\
         & Deft-news & 65.0$\genfrac{}{}{0pt}{}{+6.1}{-7.1}$ & 72.9$\genfrac{}{}{0pt}{}{+4.9}{-5.8}$ & 67.8$\genfrac{}{}{0pt}{}{+5.7}{-6.6}$ & 65.3$\genfrac{}{}{0pt}{}{+6.0}{-7.0}$\\
         STS 2014 & HDL & 58.7$\genfrac{}{}{0pt}{}{+4.5}{-4.9}$ & 63.6$\genfrac{}{}{0pt}{}{+4.1}{-4.5}$ & 61.6$\genfrac{}{}{0pt}{}{+4.3}{-4.6}$ & 60.0$\genfrac{}{}{0pt}{}{+4.4}{-4.8}$\\
         & Images & 62.4$\genfrac{}{}{0pt}{}{+4.2}{-4.6}$ & 80.9$\genfrac{}{}{0pt}{}{+2.3}{-2.6}$ & 81.4$\genfrac{}{}{0pt}{}{+2.3}{-2.6}$ & 88.2$\genfrac{}{}{0pt}{}{+1.5}{-1.7}$\\
         & OnWN & 57.7$\genfrac{}{}{0pt}{}{+4.6}{-5.0}$ & 77.3$\genfrac{}{}{0pt}{}{+2.7}{-3.1}$ & 68.6$\genfrac{}{}{0pt}{}{+3.6}{-4.0}$ & 68.1$\genfrac{}{}{0pt}{}{+3.7}{-4.0}$\\
         & Tweet-news & 53.9$\genfrac{}{}{0pt}{}{+4.9}{-5.3}$ & 75.3$\genfrac{}{}{0pt}{}{+2.9}{-3.3}$ & 74.0$\genfrac{}{}{0pt}{}{+3.1}{-3.4}$ & 69.6$\genfrac{}{}{0pt}{}{+3.5}{-3.9}$\\ \hline
         & Answers forum & 36.7$\genfrac{}{}{0pt}{}{+8.4}{-9.1}$ & 61.3$\genfrac{}{}{0pt}{}{+6.0}{-6.7}$ & 57.6$\genfrac{}{}{0pt}{}{+6.4}{-7.2}$ & 49.4$\genfrac{}{}{0pt}{}{+7.3}{-8.1}$\\
         & Answers student & 63.6$\genfrac{}{}{0pt}{}{+4.1}{-4.5}$ & 68.6$\genfrac{}{}{0pt}{}{+3.6}{-4.0}$ & 68.8$\genfrac{}{}{0pt}{}{+3.6}{-4.0}$ & 67.1$\genfrac{}{}{0pt}{}{+3.8}{-4.1}$\\
         STS 2015 & belief & 44.8$\genfrac{}{}{0pt}{}{+7.7}{-8.5}$ & 71.8$\genfrac{}{}{0pt}{}{+4.6}{-5.3}$ & 71.8$\genfrac{}{}{0pt}{}{+4.6}{-5.3}$ & 66.5$\genfrac{}{}{0pt}{}{+5.3}{-6.1}$\\
         & HDL & 66.2$\genfrac{}{}{0pt}{}{+3.8}{-4.2}$ & 69.6$\genfrac{}{}{0pt}{}{+3.5}{-3.9}$ & 70.3$\genfrac{}{}{0pt}{}{+3.4}{-3.8}$ & 67.1$\genfrac{}{}{0pt}{}{+3.8}{-4.1}$\\
         & Images & 69.1$\genfrac{}{}{0pt}{}{+3.6}{-3.9}$ & 85.5$\genfrac{}{}{0pt}{}{+1.8}{-2.1}$ & 89.2$\genfrac{}{}{0pt}{}{+1.4}{-1.6}$ & 88.1$\genfrac{}{}{0pt}{}{+1.5}{-1.7}$\\\hline
         & Answer-Answer & 40.1$\genfrac{}{}{0pt}{}{+9.8}{-10.9}$ & 62.0$\genfrac{}{}{0pt}{}{+7.0}{-8.2}$ & 53.8$\genfrac{}{}{0pt}{}{+8.2}{-9.4}$ & 47.8$\genfrac{}{}{0pt}{}{+9.0}{-10.1}$\\
         & HDL & 61.4$\genfrac{}{}{0pt}{}{+7.2}{-8.4}$ & 68.8$\genfrac{}{}{0pt}{}{+6.0}{-7.2}$ & 70.0$\genfrac{}{}{0pt}{}{+5.8}{-6.9}$ & 65.4$\genfrac{}{}{0pt}{}{+6.6}{-7.7}$\\
         STS 2016 & Plagiarism & 54.6$\genfrac{}{}{0pt}{}{+8.5}{-9.8}$ & 80.8$\genfrac{}{}{0pt}{}{+4.1}{-5.0}$ & 78.6$\genfrac{}{}{0pt}{}{+4.5}{-5.5}$ & 75.8$\genfrac{}{}{0pt}{}{+5.0}{-6.1}$\\
         & Postediting & 53.9$\genfrac{}{}{0pt}{}{+8.3}{-9.6}$ & 82.3$\genfrac{}{}{0pt}{}{+3.7}{-4.5}$ & 84.4$\genfrac{}{}{0pt}{}{+3.3}{-4.0}$ & 79.7$\genfrac{}{}{0pt}{}{+4.2}{-5.1}$ \\
         & Question-Question & 47.2$\genfrac{}{}{0pt}{}{+9.9}{-11.3}$ & 63.3$\genfrac{}{}{0pt}{}{+7.5}{-8.9}$ & 40.5$\genfrac{}{}{0pt}{}{+10.8}{-12.0}$ & 49.9$\genfrac{}{}{0pt}{}{+9.5}{-10.9}$\\ \hline
         STS-B & STS 12-16 & 64.7$\genfrac{}{}{0pt}{}{+3.0}{-3.2}$ & 75.7$\genfrac{}{}{0pt}{}{+2.2}{-2.3}$ & 72.2$\genfrac{}{}{0pt}{}{+2.4}{-2.6}$ & 70.7$\genfrac{}{}{0pt}{}{+2.5}{-2.7}$\\ \hline
         SICK & Relatedness & 79.9$\genfrac{}{}{0pt}{}{+1.0}{-1.0}$ & 86.2$\genfrac{}{}{0pt}{}{+0.7}{-0.7}$ & 81.5$\genfrac{}{}{0pt}{}{+0.9}{-1.0}$ & 82.7$\genfrac{}{}{0pt}{}{+0.9}{-0.9}$\\\hline\hline
        \end{tabular}}
    \end{minipage}
    \label{Semantic_results}
\end{table}
\end{document}